\documentclass[letterpaper, 10 pt, conference]{ieeeconf}  

\IEEEoverridecommandlockouts                              

\usepackage{amsfonts}
\usepackage{amsmath} 
\usepackage{amssymb}  
\usepackage{graphicx}
\usepackage{subcaption}
\usepackage{float}
\usepackage{booktabs}
\usepackage{multirow}
\usepackage{xcolor}
\usepackage{hyperref}

\let\labelindent\relax

\usepackage{enumitem}

\title{\LARGE \bf
Learning Sequential Kinematic Models from Demonstrations for Multi-Jointed Articulated Objects 
}

\author{Anmol Gupta$^{1}$, Weiwei Gu$^{1}$, Omkar Patil$^{1}$, Jun Ki Lee$^{2}$, Nakul Gopalan$^{1}$
\thanks{$^{1}$School of Computation and AI, ASU, Tempe
        {\tt\small \{anmolgupta, weiweigu, opatil3, ng\}}@asu.edu}%
\thanks{$^{2}$AI Institute, Seoul National University, Seoul, South Korea
        {\tt\small junkilee@snu.ac.kr}}%
}

\begin{document}

\maketitle
\thispagestyle{empty}
\pagestyle{empty}


\begin{abstract}


As robots become more generalized and deployed in diverse environments, they must interact with complex objects—many with multiple independent joints or degrees of freedom (DoF) requiring precise control. A common strategy is object modeling, where compact state-space models are learned from real-world observations and paired with classical planning. However, existing methods often rely on prior knowledge or focus on single-DoF objects, limiting their applicability. They also fail to handle occluded joints and ignore the manipulation sequences needed to access them.
We address this by learning object models from human demonstrations. We introduce \textit{Object Kinematic Sequence Machines} (OKSMs), a novel representation capturing both kinematic constraints and manipulation order for multi-DoF objects. To estimate these models from point-cloud data, we present \textit{Pokenet}, a deep neural network trained on human demonstrations. We validate our approach on $8{,}000$ simulated and $1{,}600$ real-world annotated samples. Pokenet improves joint axis and state estimation by over $20\%$ on real-world data compared to prior methods. Finally, we demonstrate OKSMs on a Sawyer robot using inverse kinematics-based planning to manipulate multi-DoF objects.

\end{abstract}


\section{INTRODUCTION}


\begin{figure*}[t]
\includegraphics[width=\textwidth]{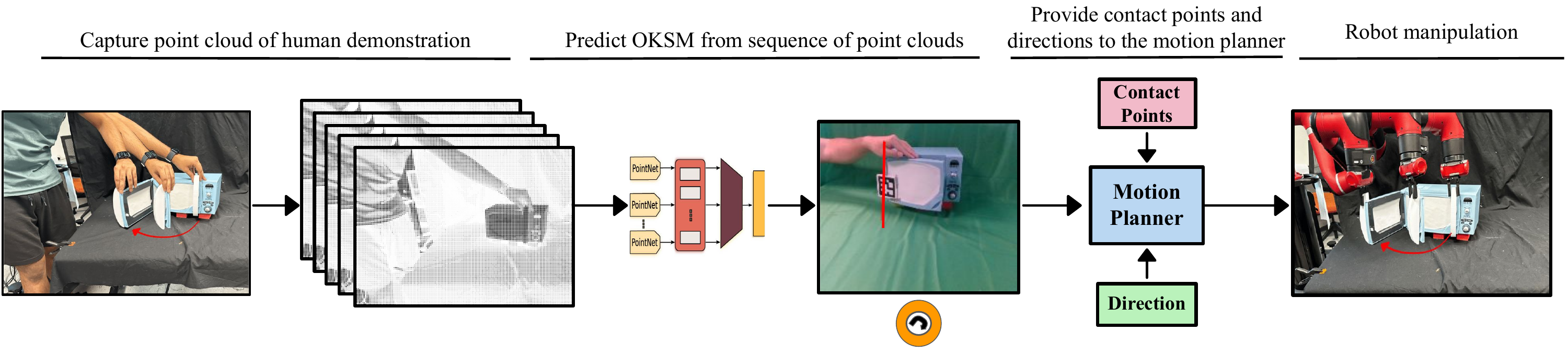}
\centering
\caption{Figure illustrating our framework enabling the robot to manipulate a microwave from a single human demonstration. We first capture the human demonstration as a sequence of point clouds. Pokenet takes this sequence and predicts an OKSM for the object. The motion planner then generates a manipulation plan for the object using the OKSM's sequence ordering and object parameterization as predicted by our model along a specified direction, such as ``open'' and a given grasp point.}
\label{fig:architecture}
\end{figure*}

Object manipulation is essential for robots in household and office environments. To perform tasks like cleaning or heating, robots must operate articulated objects such as washing machines and microwaves. They perceive the world via point clouds or 3D images and must generate continuous actions for manipulation—challenging due to noisy inputs, continuous control, and long horizons.
A promising approach is object modeling, where compact state-space representations are learned from visual data. These models enable classical planning and generalize across object instances within the same category.

Previous works on object modeling often assume prior knowledge of object categories or degrees of freedom~\cite{Abbatematteo2019LearningTG, jain2021screwnet, jain2022distributional}, limiting generalization. Others rely on hand-crafted visual features~\cite{pillai2015learning, sturm_2011}, making them sensitive to appearance variations. Flow-based approaches~\cite{zhang2024flowbot, EisnerZhang2022FLOW} aim to predict link motion but perform poorly on multi-DoF or symmetric objects. Interactive methods~\cite{4543220, Katz-2013-7694, 7487714, nie2022sfa} depend on textured objects and simple actions, limiting applicability in unstructured environments.

To overcome these challenges, we propose a category-agnostic framework that enables robots to manipulate unseen multi-DoF objects using human demonstrations. It comprises two key components: the \textit{Object Kinematic Sequence Machine} (OKSM), which encodes joint types, parameters, and manipulation order; and \textit{Pokenet}, a deep network that estimates OKSMs from demonstration point clouds.

In summary, our contributions are as follows.

\begin{figure*}[h!]
    \centering
    \begin{subfigure}{0.24\textwidth}
        \centering
        \includegraphics[width=\columnwidth, page=1]{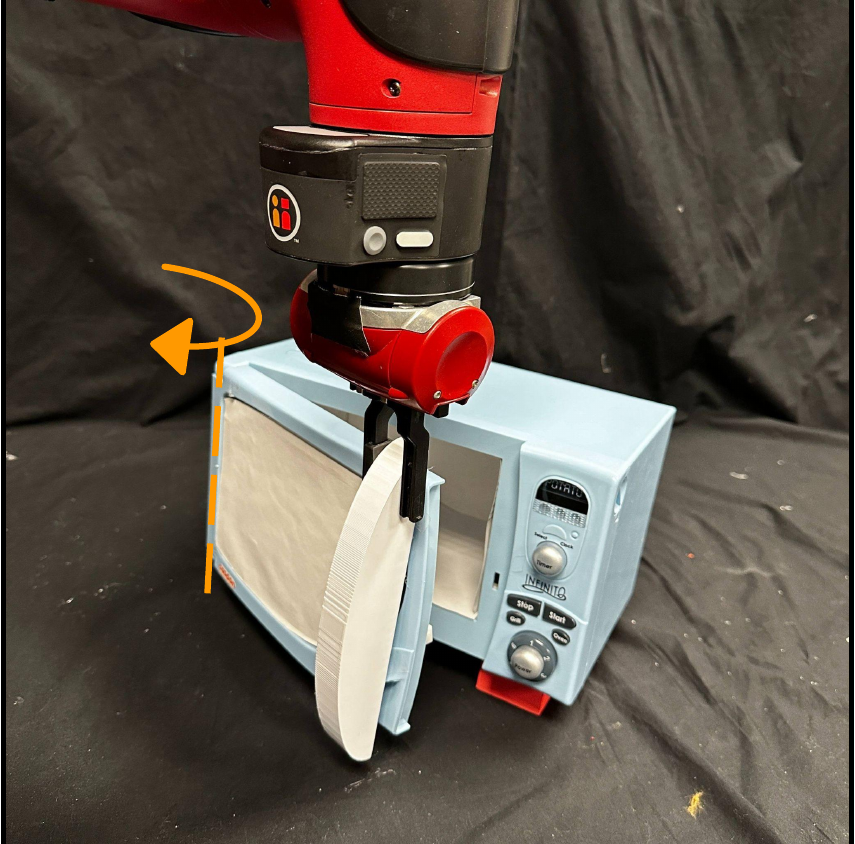}
        \caption{}
        \label{fig:intro-microwave}
    \end{subfigure}
    \begin{subfigure}{0.24\textwidth}
        \centering
        \includegraphics[width=\columnwidth, page=2]{figures/intro/intro.pdf}
        \caption{}
        \label{fig:intro-drawer}
    \end{subfigure}
    \begin{subfigure}{0.24\textwidth}
        \centering
        \includegraphics[width=\columnwidth, page=3]{figures/intro/intro.pdf}
        \caption{}
        \label{fig:intro-fridge}
    \end{subfigure}
    \begin{subfigure}{0.24\textwidth}
        \centering
        \includegraphics[width=\columnwidth, page=4]{figures/intro/intro.pdf}
        \caption{}
        \label{fig:intro-dishwasher}
    \end{subfigure}
    
    \caption{This figure shows the robot manipulating four real-world test objects using OKSMs predicted by Pokenet. (a) Microwave with a single revolute joint. (b) Drawer with a prismatic joint. (c) Fridge with two revolute and one prismatic joint. (d) Dishwasher with one revolute and one prismatic joint.
}
    \label{fig:intro-demo}
\end{figure*}

\begin{itemize}[leftmargin=*]
    \item We propose a method to represent and learn articulated objects with multiple DoFs, capturing joint parameters and manipulation sequences. To our knowledge, this is the first approach that learns object structure from human demonstrations and uses it with standard motion planners for safe and generalizable manipulation while preserving manipulation order.

    \item We introduce a real-world dataset of $5{,}500$ human-object interaction samples across seven objects from four categories, featuring varied joint types and DoFs. It is the largest annotated corpus of articulated objects to date.

    \item Experiments on simulated and real-world multi-DoF objects show that our method improves joint axis and state estimation by $30\%$ on $8{,}000$ simulated and $20\%$ on $1{,}600$ real samples. We also demonstrate generalization to unseen categories and real-world robot manipulation (Fig.~\ref{fig:intro-demo}).
\end{itemize}

\section{Related Work}

\textbf{Estimation Using Visual Data:} Prior methods estimate object properties from visual input using Gaussian mixtures~\cite{Abbatematteo2019LearningTG} or screw representations~\cite{jain2021screwnet, jain2022distributional}, but rely on object category priors and are limited to single-DoF objects. Others~\cite{li2020categorylevelarticulatedobjectpose, 10.1145/3272127.3275027} impose category-specific constraints. Pillai et al.~\cite{pillai2015learning} and Sturm et al.~\cite{sturm_2011} used hand-crafted visual features~\cite{323794, 10.1007/11744023_32}, limiting generalization. In contrast, our point-cloud-based, category-agnostic method models multi-DoF objects and infers manipulation sequences from a single human demonstration.

\textbf{Learning Articulation Flow:} Zhang et al.~\cite{EisnerZhang2022FLOW, zhang2024flowbot} predict dense motion flow from single RGBD images, but struggle with multi-DoF and symmetric objects. Our use of human demonstrations provides richer cues—such as joint limits and hidden DoFs—enabling prediction of both manipulation order and joint movement range.

\textbf{Interactive Methods:} Interactive approaches~\cite{4543220, Katz-2013-7694, 7487714, nie2022sfa} rely on textured objects and simple actions like pushing or pulling, limiting applicability. Our approach, based on human demonstrations, removes the need for texture and captures complex interactions, improving safety and generalization in real-world scenarios.

\section{Background}
Articulated objects consist of rigid links connected by joints enabling relative motion. Joints are typically revolute or prismatic, defined by an axis and position—together called articulation parameters. Prismatic joints translate along the axis; revolute joints rotate around it.

We introduce the \emph{Object Kinematic Sequence Machine} (OKSM), a structured representation encoding joint configuration and manipulation order. An OKSM is a directed chain graph $O = (N_G, E_G)$, where nodes $N_G = \{1, ..., n\}$ represent joints and edges $E_G = \{1, ..., n{-}1\}$ define the sequence. Each node includes joint type $\mathcal{M}_{i}$, axis direction $\Vec{d_i}$, position $\Vec{p_i}$, state $q_i$, and contact pose $C_i \subset SE(3)$.

This allows robots to replicate human-like sequential manipulation. OKSMs are estimated directly from point-cloud sequences, yielding compact, plannable representations that generalize across multi-DoF objects.

\section{Dataset}

For training the proposed model, we used two datasets: one simulated and one real-world.

\subsection{Simulated Dataset}
We constructed simulated dataset using PartNet-Mobility~\cite{Xiang_2020_SAPIEN, chang2015shapenet, Mo_2019_CVPR}, rendering objects and simulating realistic joint motions. Sequences of point clouds captured geometry and articulation over time—critical for multi-DoF modeling. Data was collected in the camera frame to aid generalization. We gathered $56{,}000$ samples across seven categories: microwave, laptop, washing machine, fridge, drawer, box, and furniture. Of these, $48{,}000$ were used for training and $8{,}000$ for testing, with the furniture category held out for testing due to its diversity.

\subsection{Real-World Object Dataset}
In the absence of a suitable real-world dataset, we collected one using four household objects: microwave, dishwasher, refrigerator, and drawer. These span common joint types—revolute in microwaves, and both revolute and prismatic in refrigerators and dishwashers.

Using ArUco markers and a dual-camera setup, we recorded joint parameters and motion ranges, yielding $5{,}500$ annotated samples. Of these, $3{,}900$ were used for training and $1{,}600$ for testing. To our knowledge, this is the largest annotated real-world articulated object dataset to date.

\begin{figure*}[]
    \centering
    \begin{subfigure}{0.24\textwidth}
        \centering
        \includegraphics[width=\columnwidth, page=1]{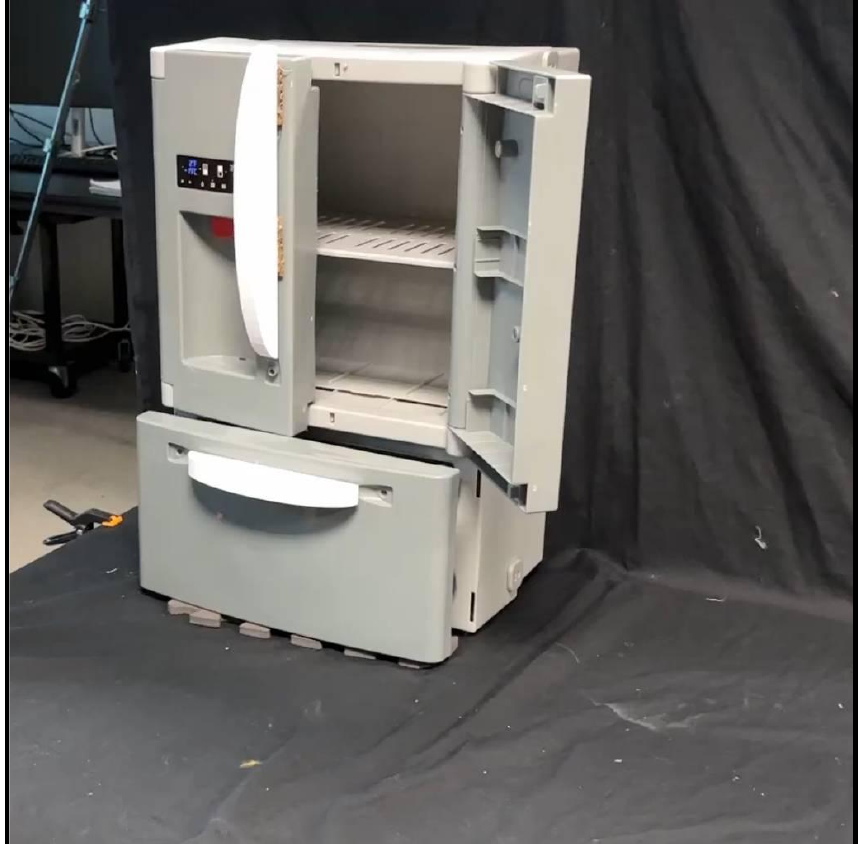}
        \caption{}
        \label{fig:sub1}
    \end{subfigure}
    \begin{subfigure}{0.24\textwidth}
        \centering
        \includegraphics[width=\columnwidth, page=2]{figures/demo/demo.pdf}
        \caption{}
        \label{fig:sub2}
    \end{subfigure}
    \begin{subfigure}{0.24\textwidth}
        \centering
        \includegraphics[width=\columnwidth, page=3]{figures/demo/demo.pdf}
        \caption{}
        \label{fig:sub3}
    \end{subfigure}
    \begin{subfigure}{0.24\textwidth}
        \centering
        \includegraphics[width=\columnwidth, page=4]{figures/demo/demo.pdf}
        \caption{}
        \label{fig:sub4}
    \end{subfigure}
    \caption{This figure shows Sawyer robot manipulating the two joints of the fridge in the order of demonstration estimated by Pokenet. (a) and (b) shows human demonstrations while (c) and (d) shows robot manipulating the object.}
    \label{fig:demo}
\end{figure*}

\section{Methods}

We propose \textit{Pokenet}, a deep learning model that estimates joint parameters and manipulation order for articulated objects from point-cloud sequences. Given link motion, it predicts joint axis position and direction, type, degrees of freedom, state at each step, and manipulation sequence.

Pokenet requires no prior knowledge of object class or DoFs, supports sequential control by predicting manipulation order, and tracks joint motion through explicit parameter estimation.

\subsection{Architecture}

The Pokenet architecture uses PointNet~\cite{qi2017pointnet} as a backbone to encode point cloud sequences. PointNet extracts features for each frame, which are passed to a transformer encoder~\cite{vaswani2017attention} to capture temporal motion across the sequence. The resulting embedding is decoded by a Multi-Layer Perceptron (MLP) to predict joint parameters: axis direction, position, angles, DoFs, joint type, and manipulation order. The entire architecture, PointNet, the transformer, and MLP, is trained from end to end on the simulated dataset. Fig.~\ref{fig:architecture} illustrates the full network.

\begin{figure}[t]
\includegraphics[width=\columnwidth]{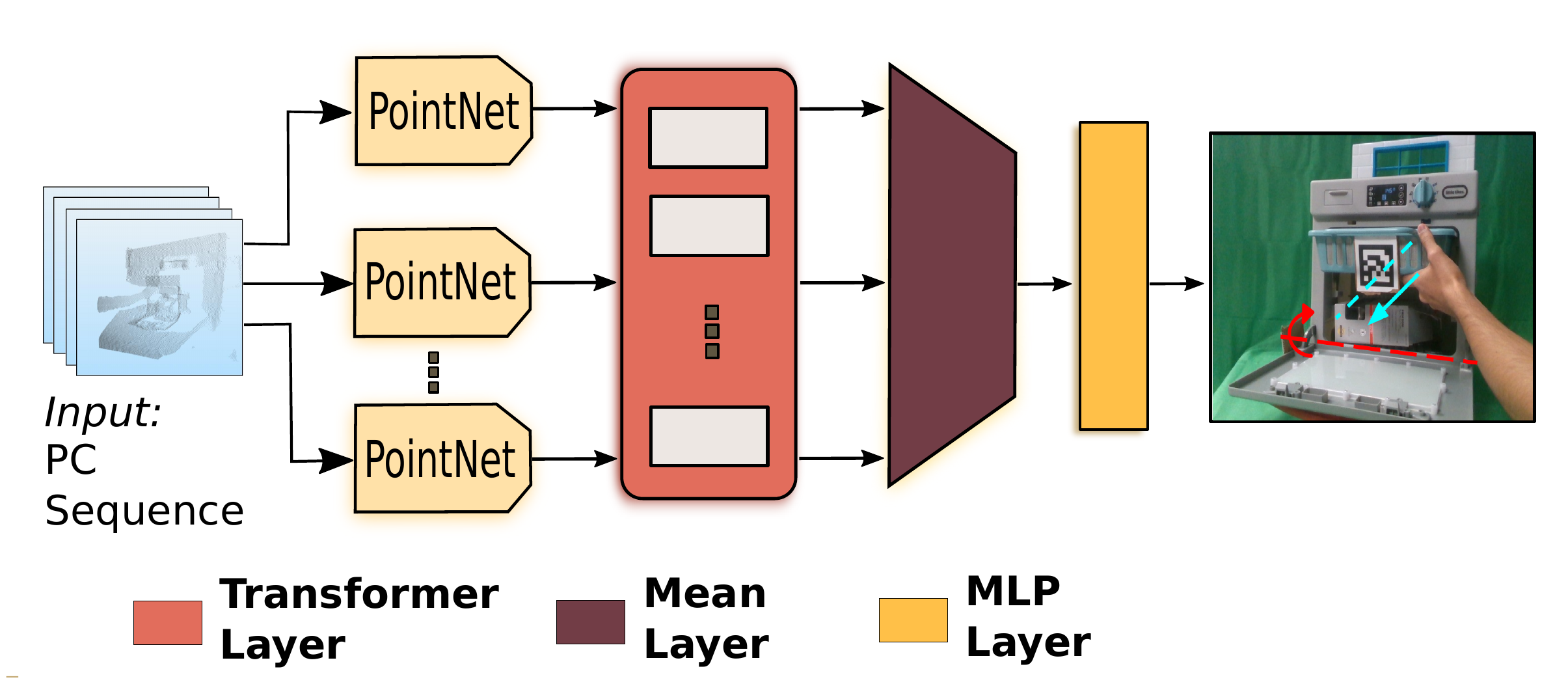}
\centering
\caption{Our model processes a sequence of point clouds, each encoded by PointNet to extract spatial features. These features are stacked and passed through a transformer encoder to capture temporal information. The encoder output is averaged and fed to an MLP to predict the OKSM.
}
\label{fig:architecture}
\end{figure}

Once the initial training on the simulated dataset is complete, we fine-tune the Pokenet model on a real-world dataset which allows the network to adapt to the real-world data.

\subsection{Loss Function}
Our model predicts joint axes and their manipulation order. Each axis consists of a direction and a position. In addition, Pokenet estimates the angle or displacement of each link over time, providing an estimate of the state of the joint.

We train the model using a compound loss function that jointly minimizes errors in direction, position, configuration, and classification tasks:

$$\mathcal{L} = \lambda_1 \mathcal{L}_{dir} + \lambda_2 \mathcal{L}_{pos} + \lambda_3 \mathcal{L}_{ord} + \lambda_4 \mathcal{L}_{dof} + \lambda_5 \mathcal{L}_{q} + \lambda_6 \mathcal{L}_{norm}$$

where $\lambda_i$ are hyperparameters.

$\mathcal{L}_{dir}$ penalizes errors in joint axis direction,  
$\mathcal{L}_{pos}$ penalizes position estimation error,  
$\mathcal{L}_{ord}$ penalizes manipulation order classification error,  
$\mathcal{L}_{dof}$ penalizes DoF classification error,  
$\mathcal{L}_{q}$ penalizes joint state (angle/displacement) error, and  
$\mathcal{L}_{norm}$ encourages unit norm for predicted direction vectors.

We set $\lambda_i = 1$ for all $i \in \{1,2,3,4,5,6\}$ based on empirical tuning.

\section{Results}
\begin{table*}[h!]
\renewcommand{\arraystretch}{1.2} 
  \centering
  \begin{tabular}{l|l|ccc|ccc}
    \toprule
    \multicolumn{1}{c|}{\multirow{2}{*}{Object}}& \multicolumn{1}{c|}{\multirow{2}{*}{Method}} & \multicolumn{3}{c|}{Axis Direction Error(Degrees)} & \multicolumn{3}{c}{Axis Position Error(Centimeters)} \\
    & &Axis 1($\downarrow$) & Axis 2($\downarrow$) & Axis 3($\downarrow$) & Axis 1($\downarrow$) & Axis 2($\downarrow$) & Axis 3($\downarrow$) \\
    \midrule
    \multirow{3}{*}{Microwave} & Screwnet* & 24.959 $\pm$ 2.236 & - & - & 15.271 $\pm$ 1.22 & - & - \\
     & GRU & {18.83 $\pm$ 0.905} & - & - & {11.1 $\pm$ 0.75} & - & - \\  
     & Ours & \textbf{15.24 $\pm$ 0.733} & - & - & \textbf{6.1 $\pm$ 0.173} & - & - \\  
    \midrule
    \multirow{3}{*}{Washing Machine} & Screwnet* & 22.571 $\pm$ 1.249 & - & - & 18.18 $\pm$ 0.857 & - & - \\
     & GRU & {15.36 $\pm$ 0.925} & - & - & {12.47 $\pm$ 0.456} & - & - \\
     & Ours & \textbf{12.994 $\pm$ 0.81} & - & - & \textbf{10.209 $\pm$ 0.722} & - & - \\
    \midrule
    \multirow{3}{*}{Laptop} & Screwnet* & 19.19 $\pm$ 0.832 & - & - & 10.33 $\pm$ 0.337 & - & - \\
    & GRU & {16.94 $\pm$ 0.695} & - & - & {5.99 $\pm$ 0.284} & - & - \\
    & Ours & \textbf{15.047 $\pm$ 0.381} & - & - & \textbf{5.272 $\pm$ 0.60} & - & - \\
    \midrule
    \multirow{3}{*}{Fridge} & Screwnet* & 21.16 $\pm$ 1.267 & 23.03 $\pm$ 0.635 & - & 20.21 $\pm$ 1.266 & 18.4 $\pm$ 0.499 & - \\
    & GRU & {16.9 $\pm$ 0.721} & {13.154 $\pm$ 0.457} & - & {12.822 $\pm$ 0.811} & {13.154 $\pm$ 0.457} & - \\
    & Ours & \textbf{15.169 $\pm$ 0.093} & \textbf{11.037 $\pm$ 0.552} & - & \textbf{12.014 $\pm$ 0.198} & \textbf{10.843 $\pm$ 0.823} & - \\
    \midrule
    \multirow{3}{*}{Drawer} & Screwnet* & 21.85 $\pm$ 1.154 & - & - & 22.431 $\pm$ 1.031 & - & - \\
    & GRU & {13.96 $\pm$ 0.462} & - & - & {13.68 $\pm$ 0.729} & - & - \\
    & Ours & \textbf{11.093 $\pm$ 0.513} & - & - & \textbf{10.792 $\pm$ 0.343} & - & - \\
    \midrule
    \multirow{3}{*}{Furniture} & Screwnet* & 27.76 $\pm$ 1.095 & 28.69 $\pm$ 1.645 & - & 19.394 $\pm$ 0.988 & 23.162 $\pm$ 1.417 & - \\
    & GRU & {18.33 $\pm$ 0.638} & {22.51 $\pm$ 1.271} & - & {12.892 $\pm$ 0.747} & {17.892 $\pm$ 0.995} & - \\
    & Ours & \textbf{17.21 $\pm$ 1.043} & \textbf{19.394 $\pm$ 0.185} & - & \textbf{10.142 $\pm$ 0.688} & \textbf{14.853 $\pm$ 0.702} & - \\
    \midrule
    \multirow{3}{*}{Box} & Screwnet* & 23.42 $\pm$ 0.926 & 22.97 $\pm$ 1.12 & - & 14.269 $\pm$ 0.517 & 14.274 $\pm$ 0.795 & - \\
    & GRU & {16.96 $\pm$ 0.753} & {15.53 $\pm$ 0.607} & - & {9.127 $\pm$ 0.396} & {10.839 $\pm$ 0.646} & - \\
    & Ours & \textbf{13.43 $\pm$ 0.220} & \textbf{13.861 $\pm$ 0.053} & - & \textbf{8.47 $\pm$ 1.043} & \textbf{8.270 $\pm$ 0.408} & - \\
    \bottomrule
  \end{tabular}
  \caption{Simulation results for seven multi-jointed objects. We present Mean Error Values for Joint Axis Directions (in Degrees) and Joint Axis Positions (in Centimeters) along with $95\%$ Confidence Interval for Partnet-Mobility Dataset. * denotes that Screwnet implemention was extended to function with Multi-Jointed Objects for Comparison.}
  \label{tab:sim-table}
\end{table*}

\begin{table*}[t]
\renewcommand{\arraystretch}{1.2} 
  \centering
  \begin{tabular}{l|l|ccc|ccc}
    \toprule
    \multicolumn{1}{c|}{\multirow{2}{*}{Object}}& \multicolumn{1}{c|}{\multirow{2}{*}{Method}} & \multicolumn{3}{c|}{Axis Direction Error(Degrees)} & \multicolumn{3}{c}{Axis Position Error(Centimeters)} \\
    & &Axis 1($\downarrow$) & Axis 2($\downarrow$) & Axis 3($\downarrow$) & Axis 1($\downarrow$) & Axis 2($\downarrow$) & Axis 3($\downarrow$) \\
    \midrule
    \multirow{3}{*}{Microwave} & Screwnet* & 30.702 $\pm$ 2.493 & - & - & 28.834 $\pm$ 1.94 & - & - \\
     & GRU & {22.047 $\pm$ 1.424} & - & - & {21.293 $\pm$ 1.65} & - & - \\
     & Ours & \textbf{19.259 $\pm$ 0.28} & - & - & \textbf{19.753 $\pm$ 0.918} & - & - \\
    \midrule
    \multirow{3}{*}{Fridge} & Screwnet* & 32.96 $\pm$ 2.077 & 31.809 $\pm$ 1.715 & 35.923 $\pm$ 2.169 & 32.193 $\pm$ 1.873 & 33.571 $\pm$ 2.267 & 33.926 $\pm$ 2.178 \\
    & GRU & {24.022 $\pm$ 1.286} & {21.173 $\pm$ 1.322} & {26.056 $\pm$ 1.462} & {24.144 $\pm$ 1.655} & {26.273 $\pm$ 1.744} & {25.337 $\pm$ 1.957} \\
    & Ours & \textbf{22.022 $\pm$ 0.281} & \textbf{19.509 $\pm$ 2.631} & \textbf{23.592 $\pm$ 2.071} & \textbf{21.40 $\pm$ 1.71} & \textbf{21.092 $\pm$ 2.24} & \textbf{23.088 $\pm$ 2.040} \\
    \midrule
    \multirow{3}{*}{Drawer} & Screwnet* & 35.284 $\pm$ 1.874 & - & - & 34.239 $\pm$ 2.135 & - & - \\
    & GRU & {23.986 $\pm$ 1.366} & - & - & {28.253 $\pm$ 1.976} & - & - \\
    & Ours & \textbf{22.147 $\pm$ 1.06} & - & - & \textbf{25.863 $\pm$ 1.742} & - & - \\
    \midrule
    \multirow{3}{*}{Dishwasher} & Screwnet* & 27.923 $\pm$ 1.016 & 33.722 $\pm$ 2.275 & - & 25.032 $\pm$ 1.782 & 37.239 $\pm$ 2.372 & - \\
    & GRU & {21.16 $\pm$ 1.144} & {27.084 $\pm$ 1.631} & - & {21.568 $\pm$ 1.612} & {30.836 $\pm$ 1.902} & - \\
    & Ours & \textbf{19.702 $\pm$ 1.34} & \textbf{24.312 $\pm$ 1.489} & - & \textbf{18.823 $\pm$ 1.209} & \textbf{27.10 $\pm$ 1.719} & - \\
    \bottomrule
  \end{tabular}
  \caption{Results on the real-world dataset with four different objects. We present Mean Error Values for Joint Axis Directions (in Degrees) and Joint Axis Positions (in Centimeters) along with $95\%$ Confidence Interval for the Real-World Dataset. * denotes that Screwnet implemention was extended to function with Multi-Jointed Objects for Comparison.}
  \label{tab:real-table}
\end{table*}

We evaluate Pokenet on simulated and real-world datasets. The model is trained jointly across all object categories without requiring category-specific knowledge.

\subsection{Simulated Dataset Results}

We first present results of Pokenet on the PartNet-Mobility simulated dataset using 12-frame point-cloud sequences as input to predict joint parameters. It was trained on $48{,}000$ samples and tested on $8{,}000$, with around eight objects per category. The ``furniture'' category was excluded from training to assess generalization.

We measured angular error in axis direction and Euclidean error in position, comparing against an extended version of Screwnet~\cite{jain2021screwnet} and a GRU-based ablation. As shown in Table~\ref{tab:sim-table}, Pokenet significantly outperforms both baselines in joint axis estimation.

\subsection{Real-World Dataset Results}

We collected $5{,}500$ real-world samples across four household objects: microwave, drawer, refrigerator, and dishwasher. ArUco markers provided ground truth for supervised training. Of these, $3{,}900$ samples were used for training and $1{,}600$ for testing.

Direction and position errors were computed as angular and Euclidean differences, respectively. Table~\ref{tab:real-table} shows Pokenet outperforms Screwnet across all object classes.

To demonstrate applicability, we used Pokenet-estimated OKSMs and provided contact points to a motion planner. For prismatic joints, the robot followed a linear trajectory (1cm steps); for revolute joints, it traced arcs (1° increments). Fig.~\ref{fig:demo} shows the robot manipulating a fridge.



\section{Conclusion}


We proposed a novel framework that learns kinematic constraints and manipulation sequences of multi-DoF objects from human demonstrations. Our approach outperforms state-of-the-art methods on both simulated and real-world datasets and includes a newly collected, annotated real-world dataset with human interactions. It makes no assumptions about object class or degrees of freedom and requires no prior object knowledge. We demonstrated successful real-world manipulation using the learned representations. Our future work will focus on detecting contact points at all joints and integrating collision-avoidance planning.










\bibliographystyle{IEEEtran}
\bibliography{ref}


\end{document}